\documentclass[a4paper]{article}

\usepackage{INTERSPEECH2019}

\title{Adapting a FrameNet Semantic Parser for Spoken Language Understanding using Adversarial Learning }
\name{Gabriel Marzinotto$^{1,2}$, G\'eraldine Damnati$^1$, Fr\'ed\'eric B\'echet$^2$}
\address{
  $^1$Orange Labs/ Lannion France\\
  $^2$Aix Marseille Universit\'e, Univ. Toulon, CNRS, LIS,	Marseille, France}
\email{gabriel.marzinotto@orange.com, geraldine.damnati@orange.com, frederic.bechet@lis-lab.fr}

\begin{document}

\maketitle
\begin{abstract}

This paper presents a new semantic frame parsing model, based on Berkeley FrameNet, adapted to process spoken documents in order to perform information extraction from broadcast contents. Building upon previous work that had shown the effectiveness of adversarial learning for domain generalization in the context of semantic parsing of encyclopedic written documents, we propose to extend this approach to elocutionary style generalization. The underlying question throughout this study is whether adversarial learning can be used to combine data from different sources and train models on a higher level of abstraction in order to increase their robustness to lexical and stylistic variations as well as automatic speech recognition errors. The proposed strategy is evaluated on a French corpus of encyclopedic written documents and a smaller corpus of radio podcast transcriptions, both annotated with a FrameNet paradigm. We show that adversarial learning increases all models generalization capabilities both on manual and automatic speech transcription as well as on encyclopedic data.



\end{abstract}

\section{Introduction}

Semantic parsing is essential for Spoken Language Understanding (SLU). Currently, the most common strategy to extract semantic information consists on a pipeline processing composed of an automatic speech recognition (ASR) system and then a semantic parser on the ASR outputs. However, most semantic parsers are built to process written language and are not robust to speech processing and even less to ASR errors. 

Even when the speech corpora to train semantic parsers is available, systems trained on perfect transcriptions tend to degrade processing ASR. A common strategy to compensate the system consists on simulating ASR errors in the textual training corpus~\cite{simonnet:hal-01715923}. Additionally, ASR systems are generally tuned measuring word error rate (WER) on a validation corpus, but this metric is not optimal for the subsequent SLU task (semantic parsing, NER, etc.). To compensate this, some specialized metrics to tune ASR have been proposed \cite{Jannet2017}. However, the number of SLU task applied on the ASR output may be large and considering dedicated metrics for each one of them is not feasible. 

Biases associated to writing, speech and ASR are a major problem in SLU task. Models learn these biases as useful information and experience a significant performance drop whenever they are applied on data from a different source. A recent approach attempting to tackle domain biases and build robust systems consists in using neural networks and adversarial learning to build domain independent representations \cite{ganin2014unsupervised}. In the NLP community, this method has been mostly used for cross-lingual transfer learning \cite{D17-1302} and more recently in monolingual setups in order to alleviate domain bias in semantic parsers~\cite{naacl-advlearning}.

In this paper we implement a FrameNet \cite{fillmore2004framenet} semantic parser, originally designed and trained to process encyclopedic texts, for semantic analysis of spoken documents (radio podcasts) in an Information Extraction perspective. We address the issue of generalization capacities of the Semantic Frame Parser when processing speech. We show that adversarial learning can be used to combine different sources of data to build robust representations and improve the generalization capacities of semantic parsers on speech and ASR data. We propose an adversarial framework based on a domain classification task that we use as a regularization technique on a state-of-the-art semantic parser. 


\section{Related Work}
\label{sec:related}

Structured semantic representations for speech processing have been mainly explored in the domain of conversational speech processing. 
Experiments around the adaptability of Framenet semantic parsing to the context of dialogues are reported on the Communicator2000 corpus \cite{Stoyanchev:2016} and in the LUNA Italian dialogue corpus \cite{coppola2009shallow}, showing their viability for labeling conversational speech. French conversational speech have also been explored on the DECODA corpus with adaptation of parsers to highly spontaneous speech with a specific adaptation process towards disfluencies and ASR errors \cite{bechet2014adapting} and  the introduction of multi-task learning to jointly handle syntactic and semantic analysis \cite{tafforeau2016joint}.
Other semantic models have also been experimented for SLU, such as Abstract Meaning Representation (AMR) in \cite{chen2016syntax} where the authors show that syntactic and semantic structured representations can help guiding attention based models neural networks. In a broader perspective, few works have been dedicated to semantic parsing of spoken contents for Information Extraction.
 
Concerning the crucial issue of model robustness, several strategies have been studied in order to improve generalization in supervised learning. A popular approach that emerged in image processing \cite{JMLR:v17:15-239} consists in training models on a double objective composed of a task-specific classifier and an adversarial domain classifier. The latter is called adversarial because it is connected to the task-specific classifier through a gradient reversal layer. During training a saddle point is searched where the task-specific classifier is good and the domain classifier is bad. It has been shown that this guarantees the resulting model to be domain independent \cite{ganin2015unsupervised}.
In Natural Language Processing tasks, this approach has been used to build cross-lingual models, doing transfer learning from English to low resource languages for POS tagging \cite{D17-1302} and sentiment analysis \cite{2016arXiv160601614C}, by using language classifiers with an adversarial objective to train task-specific but language agnostic representations. This technique is not only useful in cross-lingual transfer problems, as it has been used to improve generalization in a document classification\cite{DBLP:journals/corr/LiuQH17}, Q\&A systems \cite{yu2018modelling}, duplicate question detection \cite{Shah2018AdversarialDA} and semantic parsing \cite{naacl-advlearning} in a monolingual setup. 

In Frame Semantic Parsing, data is scarce and evaluation campaigns rarely study the generalization capacities on out-of-domain test data. Recently, the YAGS corpus was published along with the first in depth study of the domain adaptation problem in Semantic Frame Parsing\cite{hartmann2017out}. They show that the main bottleneck in domain adaptation is at the Frame Identification step and propose a more robust classifier for this task, using predicate and context embeddings to perform Frame Identification. This approach is suitable for cascade systems such as SEMAFOR \cite{das2014frame}, \cite{hermann2014semantic}. In this paper we study the generalization issue within the framework of a sequence tagging semantic frame parser that performs frame selection and argument classification in one step. And we will show that adversarial domain adaptation paradigms can be transposed into speech adaptation. 



\section{Semantic parsing model with an adversarial training scheme}
\label{sec:models}

\subsection{Semantic parsing model: \texttt{biGRU} }
\label{sec:bigru}
We implement our semantic frame parser using a sequence tagger that performs frame selection and argument classification in one step. Our model is a 4 layer bi-directional GRU tagger ($biGRU$). The advantage of this architecture is its flexibility as it can be applied on both SRL~\cite{he2017deep} and Frame Parsing~\cite{SoinFrameParsing, multijoint}. This model relies on a rich set of features including pretrained word embedding, syntactic, morphological and surface features. More details on the architecture can be found in \cite{marzinotto:hal-01731385}.

\subsection{Sequence encoding/decoding}
\label{sec:decoding}
We use a BIO label encoding in all our experiments.
On inference, we apply the coherence filter \cite{naacl-advlearning} that selects the most probable Frame for the LU and filters all incompatible FE. To ensure that output sequences respect the BIO constrains we implement an A$^*$ decoding strategy as the one proposed by \cite{he2017deep}. 

Finally, we introduce a hyper-parameter $\delta \in (-1;1)$ that is added to the output probability of the \textit{null} label $P(y_t=O)$ at each time-step. For each word, the most probable non-null hypothesis is selected if its probability is higher than $P(y_t=O)$. Varying $\delta>0$ (resp. $\delta<0$) is equivalent to being more strict (resp. less strict) on the highest non-null hypothesis. This technique allows to tune the performance of our models and study their precision/recall (P/R) trade-off. The optimal value for $\delta$ is selected on a validation set. In this paper, we either provide the P/R curve or report scores for the $Fmax$ setting.


\subsection{Adversarial Domain Classifier}
\label{sec:adv_model}

\begin{figure}[htbp]
  \centering
  \includegraphics[width=0.9\linewidth]{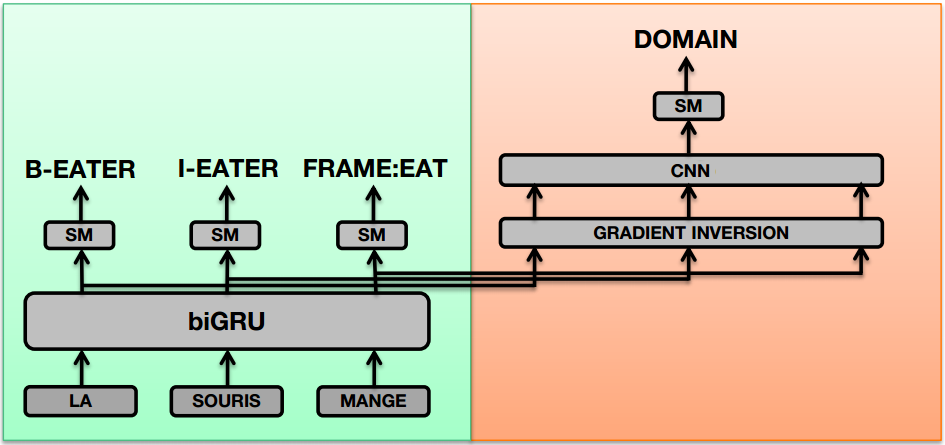}
  \caption{Adversarial Domain Classifier model}
  \label{fig:al1}
\end{figure}

Adversarial domain training was initially proposed in \cite{ganin2014unsupervised} and adapted for semantic parsing in \cite{naacl-advlearning}. We start from our $biGRU$ semantic parser and on the last hidden layer, we stack a CNN with a decision layer to implement a domain classifier (called adversarial task). The domain classifier is connected to the $biGRU$ using a gradient reversal layer. Training consists in finding a saddle point where the semantic parser is good and the domain classifier bad. This optimizes the model to be domain independent. The architecture diagram is shown in Figure~\ref{fig:al1}. 

More precisely, the adversarial classifier is trained to predict domains (\textit{i.e.} to minimize the loss $L_{adv}$) while the main task model is trained to make the adversarial task fail (\textit{i.e.} to minimize the loss $L_{frame}-L_{adv}$). In practice, in order to ensure stability during training, we follow the guidelines from ~\cite{ganin2014unsupervised}. The adversarial task gradient magnitude is attenuated by a factor $\lambda$ as shown in equation (\ref{eq:1}). Here $\nabla L$ represents the gradients w.r.t the weights $\theta$ for either the frame classifier loss or the adversarial loss, $\theta$ are the model's parameters being updated, and $\mu$ is the learning rate. This $\lambda$ factor increases on every epoch following equation (\ref{eq:2}), where $p$ is the progress, starting at 0 and increasing linearly up to 1 at the last epoch.  
\begin{equation} \label{eq:1}
\small
 \theta \leftarrow \theta - \mu * (\nabla L_{frame} - \lambda  \nabla L_{adv})
\end{equation}
\begin{equation} \label{eq:2}
\small
 \lambda = \frac{2}{1 + \exp(- 10 \cdot p)} -1 
\end{equation}





\section{Evaluation setting}
\label{sec:corpus}
Our experimental setting allows to study the differences between encyclopedic texts and speech transcriptions on the semantic parsing task. To do this, we run experiments on two corpora. First, the \texttt{CALOR} corpus \cite{marzinotto:calor}, which is a compilation of French encyclopedic documents (from Wikipedia, Vikidia and ClioTexte) with manual FrameNet~\cite{baker1998berkeley} annotations. Second, the \texttt{PODCAST} corpus is a compilation of radio podcasts from \textit{Les P'tits Bateaux} a show from France Inter broadcast. Our corpus gathers 210 sequences of a children asking a general knowledge question through a phone call followed by an answer in the form of a conversation between an expert and a journalist, each sequence is 3 to 4 minutes long. The corpus has been manually transcribed and annotated in FrameNet semantics. Statistics about the corpora are presented in Table~\ref{tab:corpus}.

 Even though both corpora are related to general knowledge, they are very different in terms of style. \texttt{CALOR} is composed of well written encyclopedic text dealing with three subjects (WW1, Archaeology and Ancient History). On the other hand, \texttt{PODCAST} contains transcriptions of a radio show addressed to children using a simpler discourse but dealing with a broader set of general knowledge topics. These corpora have been designed in the perspective of targeted Information Extraction tasks. Due to this, we used a \textit{partial parsing} policy where only 53 Frames have been annotated on the whole corpus. This allows to rapidly annotate large corpora and yields a much higher amount of occurrences per Frame (i.e. 504 in \texttt{CALOR} vs. 33 in \texttt{FrameNet}). 


\begin{table}
\small
\centering
\normalsize
\caption{Statistics of both the \texttt{CALOR} corpus of encyclopedic documents and the \texttt{PODCAST} corpus of transcriptions }
\label{tab:corpus}
\begin{tabular}{|l|c|c|c|c|}
\hline
\textbf{Corpus}  & \textbf{\# Sentence}  & \textbf{\# Frame} & \textbf{\# FE} \\ \hline
\texttt{CALOR}   & 67381     & 26725       & 57688    \\ \hline
\texttt{PODCAST} &  4233     & 2298        &  5474    \\ \hline
\end{tabular}
\vspace*{-6mm}
\end{table}

\begin{figure}[htbp]
  \centering
  \includegraphics[width=0.9\linewidth]{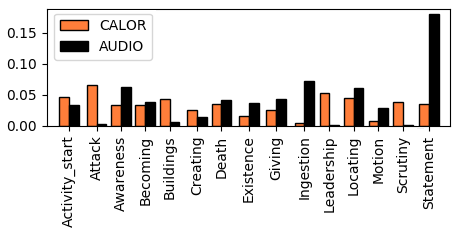}
  \caption{Most frequent frames and their normalized distribution for each partition (\texttt{CALOR} and \texttt{PODCAST})}
  \label{fig:frame_dist}
  \vspace*{-3mm}
\end{figure}

Each corpus has a different prior on the Frames and LU distributions. Figure \ref{fig:frame_dist} shows the normalized Frame distributions for both sets, illustrating the domain dependence. Frames such as  \textit{Attack}  and \textit{Leadership} are frequent in \texttt{CALOR} while \textit{Statement} and \textit{Ingestion} are common Frames in \texttt{PODCAST}. When we analyze the Lexical Units (LUs) distribution and their associated Frames, we observe an imbalance between corpora. In \texttt{PODCAST} we find that 38\% of LUs do not trigger any Frame. This is a very large number compared to the 13\% in the \texttt{CALOR} corpus. The reason for this is that in \texttt{PODCAST} many LUs appear as part of idioms or oral expressions that do not convey semantic Frames. Examples of these expressions are: \textit{c'est-\`a-dire (that is), disons (let's say), comment dire (how to say)}. Moreover, speakers in \texttt{PODCAST} tend to address children in a simplified language, using very common words which are often polysemous, this makes LUs in \texttt{PODCAST} more ambiguous. Examples for this are: \textit{\c{c}a donne} (\textit{"results in"} instead of \textit{"it gives"}), \textit{arriver \`a} (\textit{"being able to"} as opposed to \textit{"arriving"}), \textit{on se demande} (\textit{"we ask ourselves"} and not \textit{"to request"}). As a consequence, processing "simplified" language for children is not "simpler" from a semantic parsing perspective. 

To generate an ASR corpus, we process our \texttt{PODCAST} samples using the Cobalt Speech Recognition system developed at Orange Labs. It is a Kaldi-based decoder using a time-delay neural network based acoustic model \cite{povey2016purely} trained  on  more  than  2000 h  of  clean  and  noisy speech, with a  1.7 million word lexicon, and a 5-gram language model trained on 3 billion words. We further align the ASR outputs with the manual transcriptions in order to project the FrameNet annotations into the ASR corpus. The evaluation of our ASR system on the \texttt{PODCAST} corpus yields a WER of 14.2\%, with a large variation between children speech in telephone recording conditions (41\% WER) and journalist and expert studio conversation (13\% WER).

\section{Results}
\label{sec:results}

\subsection{Without Adaptation}

In these experiments we first evaluate the performances of our $biGRU$ semantic Frame parser for different training configurations in order to better understand the main difficulties of the transfer learning for SLU. For each corpus, we split data into 60\% for training 20\% for validation and 20\% for test. 
We report results using \textit{F-measure} metrics at the \textit{Frame} and \textit{Argument Identification} (respectively FI and AI) levels. Errors are cumulative: in order to obtain a correct argument identification, we need to identify the correct Frame. Since we test our model on the ASR outputs, we evaluate using a soft-span metric for AI, meaning that for an argument hypothesis to be correct the label has to match the reference but the span may not be exactly the same (an overlap with the matching reference is required). This metric does not oversimplify the argument detection task since the median length of an argument in \texttt{PODCAST} corpus is 2 words (slightly lower than for the original \texttt{CALOR} corpus where the median length is 3 words).


Results for this experience are reported in table \ref{tab:bigru_baseline}. We observe that simply learning a model on the \texttt{CALOR} textual corpus and applying it on the speech corpus yields very low performances. In \texttt{PODCAST}, for the Frame Identification phase, the  parser fails to analyse idioms and oral expressions where the LU should not trigger Frames. Training a parser on the small \texttt{PODCAST-GOLD} or \texttt{PODCAST-ASR} corpus fixes this problem increasing the FI score from 79.6\% to 86.3\% on manual transcripts and from 79.1\% to 86.3\% on automatic transcripts. However, the \texttt{PODCAST} corpus is too small to train a model for the Argument Identification task. For this task the only models that give acceptable performances are those that were trained using data from both \texttt{CALOR} and \texttt{PODCAST}. 

Even though our ASR system has a WER or 15\% when recognizing LUs, most of these errors were confusions with other inflected forms of the LU, which do not affect FI. For this reason, performances in this task are almost the same for \texttt{ASR} and \texttt{GOLD} transcripts. We cannot say the same thing from the Argument Identification task. In our  best training configuration (\texttt{CALOR+PODCAST(BOTH)}), performances on \texttt{ASR} are 5.5 points bellow the performances on \texttt{GOLD}. Even though WER inside the Arguments is lower (only 12\%), these errors deeply affect the semantic parser. There are two reasons for this: the first one is that insertions and deletions appear mostly on short words, these words are often pronouns, prepositions and articles. When a pronoun is missing a whole Argument can be lost, similarly, articles and preposition are strong indicators of the presence of a specific argument. The second reason is that  semantic roles are strongly correlated to syntax and the ASR errors easily introduce syntax errors, for example confusing \textit{"a"(to have)} with \textit{"\`a"(to)}  or  \textit{"est"(to be)} with  \textit{"et"(and)} completely change the structure and the meaning of a sentence.



\begin{table}
\centering
\caption{F-measure (Fmax) on  Frame and Argument Identification using ($biGRU$) and different training datasets.}
\label{tab:bigru_baseline}
\small
\resizebox{0.45\textwidth}{!}{%
    \begin{tabular}{lcccc}
    \cline{2-5}
      &   \multicolumn{2}{c}{Frame Ident} &  \multicolumn{2}{c}{Argument Ident} \\
     \cline{2-3}  \cline{4-5}  
      &  \multicolumn{2}{c}{PODCAST} &  \multicolumn{2}{c}{PODCAST}  \\

      &  \multicolumn{1}{c}{\texttt{GOLD}} & \multicolumn{1}{c}{\texttt{ASR}}  &  \multicolumn{1}{c}{\texttt{GOLD}} & \multicolumn{1}{c}{\texttt{ASR}}  \\ 
     \hline
      \multicolumn{1}{l}{CALOR}              & 79.6 & 79.1  & 52.3 & 51.4  \\ 
      \multicolumn{1}{l}{PODCAST$_{\rm GOLD}$}       & 86.3 & 85.0  & 55.7 & 51.0  \\ 
      \multicolumn{1}{l}{PODCAST$_{\rm ASR}$}        & 86.3 & 86.3  & 56.0 & 53.6  \\ 
      \multicolumn{1}{l}{CALOR+PODCAST$_{\rm GOLD}$} & 84.8 & 84.5  & 61.3 & 56.7  \\ 
      \multicolumn{1}{l}{CALOR+PODCAST$_{\rm ASR}$}  & 86.0 & 85.6  & 59.9 & 57.8  \\ 
      \multicolumn{1}{l}{PODCAST$_{\rm BOTH}$}       & 87.3 & 86.9  & 58.4 & 55.9  \\ 
      \multicolumn{1}{l}{CALOR+PODCAST$_{\rm BOTH}$} & 87.4 & 85.0  & 65.4 & 59.9  \\ 
     \hline
    \end{tabular}%
}
\vspace*{-6mm}
\end{table}

\subsection{With Adaptation}

 In this experiment we compare our initial $biGRU$ model with a model trained using adversarial domain adaptation. We ran experiments using some of the different configurations of the training corpus presented in Table \ref{tab:bigru_baseline}. However, since the adaptation technique behaves similarly in all these configurations, we present the results for the most common setup, which consists in training a model on all data sources (\texttt{CALOR} and \texttt{PODCAST(BOTH)}). Under this setup we trained our adversarial model ($biGRU$+$adv$) with a "domain" classifier that distinguishes between two sources, determining if samples come either from \texttt{CALOR} or \texttt{PODCAST(BOTH)}. In earlier experiments we tested different domain classification tasks varying the classes, and using unsupervised inferred domains, but the simple 2 domain task yields the best results on \texttt{PODCAST-ASR}.

\begin{figure}[htbp]
  \centering
  \includegraphics[width=1.0\linewidth]{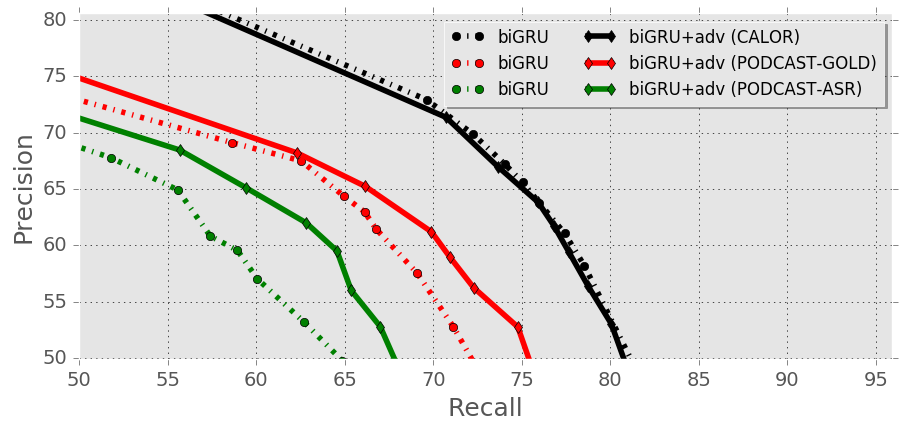}
  \caption{Precision/Recall trade-off with ($biGRU$+$adv$) and without ($biGRU$) adversarial training for the different test sets}
  \label{fig:pr_curve}
  \vspace*{-4mm}
\end{figure}

 
Results are given in figure~\ref{fig:pr_curve} where the precision/recall curve on argument identification is obtained by varying a threshold over final argument detection in two conditions: with ($biGRU$+$adv$) and without ($biGRU$) adversarial training for the  \texttt{PODCAST-GOLD} and \texttt{PODCAST-ASR} tests sets. For the sake of comparison, the results are also provided over the initial \texttt{CALOR} test set, showing that these results are not harmed by the adaptation process. Table~\ref{tab:frame_audio_calor} presents F-measure (F-max) for both Frame and Argument Identification tasks on each test corpus. When applying our adversarial method, we clearly increase the generalization capabilities of our model on both test sets, as the $biGRU$+$adv$ curve outperforms the $biGRU$ curve at every operating point in figure~\ref{fig:pr_curve}. This is confirmed on the F-max values in table~\ref{tab:frame_audio_calor}. The corpus with the highest improvements is the ASR corpus, with +2.5 points. This shows that our approach can help building higher level representation that are more independent from data source (written or spoken language) especially when dealing with transcription errors.

\begin{table}
\centering

\caption{F-measure (Fmax) on FI and AI with ($biGRU$+$adv$) and without ($biGRU$) adversarial training. }
\label{tab:frame_audio_calor}

\vspace*{-2mm}
\small
\begin{tabular}{@{\extracolsep{4pt}}lcccc@{}}
\cline{2-5}
  &  \multicolumn{2}{c}{Frame Ident.} &  \multicolumn{2}{c}{Argument Ident.} \\
 \cline{2-3}  \cline{4-5} 
  & \multicolumn{2}{c}{PODCAST} &  \multicolumn{2}{c}{PODCAST}  \\
  \cline{2-3}  \cline{4-5}   
  &  \multicolumn{1}{c}{\texttt{GOLD}} & \multicolumn{1}{c}{\texttt{ASR}} &   \multicolumn{1}{c}{\texttt{GOLD}} & \multicolumn{1}{c}{\texttt{ASR}}   \\ 
 \hline
  \multicolumn{1}{l}{$biGRU$}     & 87.4 & 85.0  & 65.4 & 59.9  \\ 
  \multicolumn{1}{l}{$biGRU$+$adv$} & 89.3 & 88.3  & 65.9 & 62.4  \\ 
 \hline
\end{tabular}
\normalsize
\vspace*{-2mm}
\end{table}

\subsection{Error Analysis}


We observed that most of the errors in \textbf{Frame Identification} are associated to LUs that are polysemous in a general context, but are unambiguous given a thematic domain. 
For example, \textit{dire (to say)} triggers the frame \textit{Statement} in most of the \texttt{CALOR} corpus, but in the \texttt{PODCAST} corpus most of the time it does not trigger any Frame as it is used as part of an oral expression. Under these circumstances, the model underestimates the FI task and assigns a Frame to a LU without doing a proper analysis of its meaning using appropriated features. In Table~\ref{tab:LU_FrameID} we show the FI score for the LUs suffering important changes in their meaning distribution across corpora. These changes are mostly due to LUs used in oral expression, rather than to a thematic change. Adversarial training is beneficial for these LUs, specially for the ASR. This result supports the idea that adaptation can be useful even if the spoken and written thematic domains are similar.

\begin{table}
\centering
\normalsize
\caption{Frame Identification score for LUs with the highest variation of sense distribution across corpora }
\label{tab:LU_FrameID}
\vspace*{-2mm}
\small
\begin{tabular}{lccccc}
\hline
LU & \multicolumn{2}{c}{GOLD} & \multicolumn{2}{c}{ASR} \\
 & $biGRU$ & +$adv$ & $biGRU$ & +$adv$ \\
\hline
\texttt{dire}     & 80.9 & 81.8 & 77.1 & 85.4 \\ 
\texttt{donner}   & 78.8 & 84.8 & 77.4 & 87.1 \\ 
\texttt{demander} & 75.8 & 75.8 & 56.0 & 72.0 \\ 
\hline
\end{tabular}
\vspace*{-2mm}
\end{table}


We focus now on the \textbf{Argument Identification} (AI) level. We try to identify precisely in which situations the adversarial learning strategy improves AI. To do so, we focus on the \texttt{PODCAST-ASR} test data and we evaluate our strategy on  the scope of some complexity factors \cite{marzinotto:hal-01731385}. As we can see in table~\ref{tab:ArgIDperf}, adversarial training largely improves the identification of non-core Frame Elements (FE) such as \textit{Place} or \textit{Time}, while having a moderate impact on core FEs (specific arguments with typical agent and patient semantic roles). This is not surprising since non-core FEs are often shared across several Frames despite having non trivial lexical variations for each Frame/domain. Consider the non-core FE \textit{Place} whose content can vary from names of countries and cities to common nouns (\textit{"in the park"}) and adverbs (\textit{"there"}). Under these circumstances adversarial learning can successfully build a higher level representation of the FE. As for the other complexity factors, bigger gains are observed for the \textit{difficult} conditions (i.e. nominal triggers and long sentences). On the other hand, adversarial learning slightly harmed performances on the short sentences.



\begin{table}
\centering
\normalsize
\caption{Argument Identification results on the \texttt{PODCAST-ASR} corpus according to complexity factors (Fmax)}
\label{tab:ArgIDperf}
\vspace*{-2mm}
\small
\begin{tabular}{lcc}
\hline
\texttt{D3}   & $biGRU$ & $biGRU$+$adv$ \\
\hline
overall          & 59.9 & 62.4 (+2.5\%)\\ 
\hline
core FE          & 63.6 & 65.5 (+1.9\%)\\ 
non-core FE      & 37.8 & 43.0 (+5.2\%)\\ 
\hline
verbal trigger   & 61.2 & 63.8 (+2.6\%)\\ 
nominal trigger  & 36.5 & 41.0 (+4.5\%)\\ 
\hline
short sentences & 65.3 & 64.8 (-0.5\%) \\ 
long sentences  & 58.8 & 61.9 (+3.1\%) \\ 
\hline
\end{tabular}
\vspace*{-6mm}
\end{table}

\begin{table}[htbp]
\centering
\normalsize
\caption{Frame Element Identification results (Fmax) on the \texttt{PODCAST-ASR} corpus under different WER }
\label{tab:wer}
\vspace*{-2mm}
\small
\begin{tabular}{lccc}
\hline
\texttt{WER}    & \#frames & $biGRU$ & $biGRU$+$adv$ \\
\hline
overall         & 489 & 59.9 & 62.4 (\%)\\ 
\hline
$0\leq WER<5$   & 50  & 70.1 & 72.5 (+2.4\%)\\ 
$5\leq WER<10$  & 167 & 63.6 & 65.4 (+1.8\%)\\ 
$10\leq WER<15$ & 126 & 61.1 & 63.8 (+2.7\%)\\ 
$15\leq WER<20$ & 81  & 55.0 & 60.7 (+5.7\%)\\ 
$20\leq WER$    & 65 & 51.8 & 58.0 (+6.2\%)\\ 
\hline
\end{tabular}
\vspace*{-2mm}
\end{table}


Finally, intuition says a higher WER translates into lower performance for our Frame parser. To corroborate this, Table \ref{tab:wer} shows the performance of our models on different subsets of \texttt{PODCAST-ASR} presenting different WER. We observe that indeed higher WER yields lower performance for the semantic parsing task. Table \ref{tab:wer} also shows that our adversarial learning strategy is more beneficial for transcriptions with a high WER.

\section{Conclusions}
We have presented a study on the robustness of Frame semantic parsing under changes in the elocutionary style. We presented an adaptation technique based on adversarial learning. This technique combines data from different sources (speech and text) to train more robust models that perform semantic parsing on a higher level of abstraction. Results  showed that domain adversarial training can be effectively used to improve the generalization capacities of our semantic frame parser on spoken documents. This positive result suggests that our approach could apply successfully to more Spoken Language Understanding tasks.

\bibliographystyle{IEEEtran}

\bibliography{mybib}

\begin{thebibliography}{10}
\providecommand{\url}[1]{#1}
\csname url@samestyle\endcsname
\providecommand{\newblock}{\relax}
\providecommand{\bibinfo}[2]{#2}
\providecommand{\BIBentrySTDinterwordspacing}{\spaceskip=0pt\relax}
\providecommand{\BIBentryALTinterwordstretchfactor}{4}
\providecommand{\BIBentryALTinterwordspacing}{\spaceskip=\fontdimen2\font plus
\BIBentryALTinterwordstretchfactor\fontdimen3\font minus
  \fontdimen4\font\relax}
\providecommand{\BIBforeignlanguage}[2]{{%
\expandafter\ifx\csname l@#1\endcsname\relax
\typeout{** WARNING: IEEEtran.bst: No hyphenation pattern has been}%
\typeout{** loaded for the language `#1'. Using the pattern for}%
\typeout{** the default language instead.}%
\else
\language=\csname l@#1\endcsname
\fi
#2}}
\providecommand{\BIBdecl}{\relax}
\BIBdecl

\bibitem{simonnet:hal-01715923}
\BIBentryALTinterwordspacing
E.~Simonnet, S.~Ghannay, N.~Camelin, and Y.~Est{\`e}ve, ``{Simulating ASR
  errors for training SLU systems},'' in \emph{{LREC 2018}}, Miyazaki, Japan,
  May 2018. [Online]. Available:
  \url{https://hal-univ-lemans.archives-ouvertes.fr/hal-01715923}
\BIBentrySTDinterwordspacing

\bibitem{Jannet2017}
\BIBentryALTinterwordspacing
M.~A.~B. Jannet, O.~Galibert, M.~Adda-Decker, and S.~Rosset, ``Investigating
  the effect of asr tuning on named entity recognition,'' in \emph{Proc.
  Interspeech 2017}, 2017, pp. 2486--2490. [Online]. Available:
  \url{http://dx.doi.org/10.21437/Interspeech.2017-1482}
\BIBentrySTDinterwordspacing

\bibitem{ganin2014unsupervised}
Y.~Ganin and V.~Lempitsky, ``Unsupervised domain adaptation by
  backpropagation,'' \emph{arXiv preprint arXiv:1409.7495}, 2014.

\bibitem{D17-1302}
\BIBentryALTinterwordspacing
J.-K. Kim, Y.-B. Kim, R.~Sarikaya, and E.~Fosler-Lussier, ``Cross-lingual
  transfer learning for pos tagging without cross-lingual resources,'' in
  \emph{Proceedings of the 2017 Conference on Empirical Methods in Natural
  Language Processing}.\hskip 1em plus 0.5em minus 0.4em\relax Association for
  Computational Linguistics, 2017, pp. 2832--2838. [Online]. Available:
  \url{http://aclweb.org/anthology/D17-1302}
\BIBentrySTDinterwordspacing

\bibitem{naacl-advlearning}
G.~Marzinotto, G.~Damnati, F.~B{\'e}chet, and B.~Favre, ``Robust semantic
  parsing with adversarial learning for domain generalization,'' in \emph{Proc.
  of NAACL}, 2019.

\bibitem{fillmore2004framenet}
C.~J. Fillmore, C.~F. Baker, and H.~Sato, ``Framenet as a "net".'' in
  \emph{LREC}, 2004.

\bibitem{Stoyanchev:2016}
\BIBentryALTinterwordspacing
S.~Stoyanchev, A.~Stent, and S.~Bangalore, ``Evaluation of semantic dependency
  labeling across domains,'' in \emph{Proceedings of the Thirtieth AAAI
  Conference on Artificial Intelligence}, ser. AAAI'16.\hskip 1em plus 0.5em
  minus 0.4em\relax AAAI Press, 2016, pp. 2814--2820. [Online]. Available:
  \url{http://dl.acm.org/citation.cfm?id=3016100.3016295}
\BIBentrySTDinterwordspacing

\bibitem{coppola2009shallow}
B.~Coppola, A.~Moschitti, and G.~Riccardi, ``Shallow semantic parsing for
  spoken language understanding,'' in \emph{Proceedings of Human Language
  Technologies: The 2009 Annual Conference of the North American Chapter of the
  Association for Computational Linguistics, Companion Volume: Short
  Papers}.\hskip 1em plus 0.5em minus 0.4em\relax Association for Computational
  Linguistics, 2009, pp. 85--88.

\bibitem{bechet2014adapting}
F.~Bechet, A.~Nasr, and B.~Favre, ``Adapting dependency parsing to spontaneous
  speech for open domain spoken language understanding,'' in \emph{Fifteenth
  Annual Conference of the International Speech Communication Association},
  2014.

\bibitem{tafforeau2016joint}
J.~Tafforeau, F.~Bechet, T.~Artieres, and B.~Favre, ``Joint syntactic and
  semantic analysis with a multitask deep learning framework for spoken
  language understanding.'' in \emph{Interspeech}, 2016, pp. 3260--3264.

\bibitem{chen2016syntax}
Y.-N. Chen, D.~Hakanni-T{\"u}r, G.~Tur, A.~Celikyilmaz, J.~Guo, and L.~Deng,
  ``Syntax or semantics? knowledge-guided joint semantic frame parsing,'' in
  \emph{2016 IEEE Spoken Language Technology Workshop (SLT)}.\hskip 1em plus
  0.5em minus 0.4em\relax IEEE, 2016, pp. 348--355.

\bibitem{JMLR:v17:15-239}
\BIBentryALTinterwordspacing
Y.~Ganin, E.~Ustinova, H.~Ajakan, P.~Germain, H.~Larochelle, F.~Laviolette,
  M.~Marchand, and V.~Lempitsky, ``Domain-adversarial training of neural
  networks,'' \emph{Journal of Machine Learning Research}, vol.~17, no.~59, pp.
  1--35, 2016. [Online]. Available:
  \url{http://jmlr.org/papers/v17/15-239.html}
\BIBentrySTDinterwordspacing

\bibitem{ganin2015unsupervised}
Y.~Ganin and V.~Lempitsky, ``Unsupervised domain adaptation by
  backpropagation,'' in \emph{International Conference on Machine Learning},
  2015, pp. 1180--1189.

\bibitem{2016arXiv160601614C}
\BIBentryALTinterwordspacing
X.~{Chen}, Y.~{Sun}, B.~{Athiwaratkun}, C.~{Cardie}, and K.~{Weinberger},
  ``Adversarial deep averaging networks for cross-lingual sentiment
  classification,'' \emph{ArXiv e-prints}, Jun. 2016. [Online]. Available:
  \url{https://arxiv.org/abs/1606.01614}
\BIBentrySTDinterwordspacing

\bibitem{DBLP:journals/corr/LiuQH17}
\BIBentryALTinterwordspacing
P.~Liu, X.~Qiu, and X.~Huang, ``Adversarial multi-task learning for text
  classification,'' \emph{CoRR}, vol. abs/1704.05742, 2017. [Online].
  Available: \url{http://arxiv.org/abs/1704.05742}
\BIBentrySTDinterwordspacing

\bibitem{yu2018modelling}
J.~Yu, M.~Qiu, J.~Jiang, J.~Huang, S.~Song, W.~Chu, and H.~Chen, ``Modelling
  domain relationships for transfer learning on retrieval-based question
  answering systems in e-commerce,'' in \emph{Proceedings of the Eleventh ACM
  International Conference on Web Search and Data Mining}.\hskip 1em plus 0.5em
  minus 0.4em\relax ACM, 2018, pp. 682--690.

\bibitem{Shah2018AdversarialDA}
D.~J. Shah, T.~Lei, A.~Moschitti, S.~Romeo, and P.~Nakov, ``Adversarial domain
  adaptation for duplicate question detection,'' in \emph{EMNLP}, 2018.

\bibitem{hartmann2017out}
S.~Hartmann, I.~Kuznetsov, T.~Martin, and I.~Gurevych, ``Out-of-domain framenet
  semantic role labeling,'' in \emph{Proceedings of the 15th Conference of the
  European Chapter of the Association for Computational Linguistics: Volume 1,
  Long Papers}, vol.~1, 2017, pp. 471--482.

\bibitem{das2014frame}
D.~Das, D.~Chen, A.~F. Martins, N.~Schneider, and N.~A. Smith, ``Frame-semantic
  parsing,'' \emph{Computational linguistics}, vol.~40, no.~1, pp. 9--56, 2014.

\bibitem{hermann2014semantic}
K.~M. Hermann, D.~Das, J.~Weston, and K.~Ganchev, ``Semantic frame
  identification with distributed word representations.'' in \emph{ACL (1)},
  2014, pp. 1448--1458.

\bibitem{he2017deep}
L.~He, K.~Lee, M.~Lewis, and L.~Zettlemoyer, ``Deep semantic role labeling:
  What works and what's next,'' in \emph{Proceedings of the Annual Meeting of
  the Association for Computational Linguistics}, 2017.

\bibitem{SoinFrameParsing}
\BIBentryALTinterwordspacing
B.~Yang and T.~Mitchell, ``A joint sequential and relational model for
  frame-semantic parsing,'' in \emph{Proceedings of the 2017 Conference on
  Empirical Methods in Natural Language Processing}.\hskip 1em plus 0.5em minus
  0.4em\relax Association for Computational Linguistics, 2017, pp. 1247--1256.
  [Online]. Available: \url{http://aclweb.org/anthology/D17-1128}
\BIBentrySTDinterwordspacing

\bibitem{multijoint}
\BIBentryALTinterwordspacing
A.~Celikyilmaz, , J.~Gao, , and Y.-Y. Wang, ``Multi-domain joint semantic frame
  parsing using bi-directional rnn-lstm.''\hskip 1em plus 0.5em minus
  0.4em\relax ISCA, June 2016. [Online]. Available:
  \url{https://www.microsoft.com/en-us/research/publication/multijoint/}
\BIBentrySTDinterwordspacing

\bibitem{marzinotto:hal-01731385}
\BIBentryALTinterwordspacing
G.~Marzinotto, F.~B{\'e}chet, G.~Damnati, and A.~Nasr, ``{Sources of Complexity
  in Semantic Frame Parsing for Information Extraction},'' in \emph{{
  International FrameNet Workshop 2018}}, Miyazaki, Japan, May 2018. [Online].
  Available: \url{https://hal.archives-ouvertes.fr/hal-01731385}
\BIBentrySTDinterwordspacing

\bibitem{marzinotto:calor}
\BIBentryALTinterwordspacing
G.~Marzinotto, J.~Auguste, F.~B{\'e}chet, G.~Damnati, and A.~Nasr, ``{Semantic
  Frame Parsing for Information Extraction : the CALOR corpus},'' in \emph{{
  LREC 2018}}, Miyazaki, Japan, May 2018. [Online]. Available:
  \url{https://hal.archives-ouvertes.fr/hal-01959187}
\BIBentrySTDinterwordspacing

\bibitem{baker1998berkeley}
C.~F. Baker, C.~J. Fillmore, and J.~B. Lowe, ``The berkeley framenet project,''
  in \emph{Proceedings of the 36th Annual Meeting of the Association for
  Computational Linguistics and 17th International Conference on Computational
  Linguistics-Volume 1}.\hskip 1em plus 0.5em minus 0.4em\relax Association for
  Computational Linguistics, 1998, pp. 86--90.

\bibitem{povey2016purely}
D.~Povey, V.~Peddinti, D.~Galvez, P.~Ghahremani, V.~Manohar, X.~Na, Y.~Wang,
  and S.~Khudanpur, ``Purely sequence-trained neural networks for asr based on
  lattice-free mmi.'' in \emph{Interspeech}, 2016, pp. 2751--2755.

\end{thebibliography}

\end{document}